\def\year{2019}\relax
\documentclass[letterpaper]{article} 
\usepackage{aaai19}  
\usepackage{times}  
\usepackage{helvet}  
\usepackage{courier}  
\usepackage{url}  
\usepackage{graphicx}  

\usepackage{amssymb,amsmath}
\usepackage{algorithmic,algorithm}
\usepackage{relsize}
\usepackage{mathtools}
\usepackage{bm}
\usepackage{algorithmic,algorithm}
\usepackage{color}

\usepackage{subcaption}
\usepackage{booktabs}

\usepackage{bm} 

\title{Bayesian graph convolutional neural networks for semi-supervised classification}
\author{Yingxue Zhang\thanks{These authors contributed equally to this
  work.}\\
Huawei Noah's Ark Lab\\
Montreal Research Centre\\
7101 Avenue du Parc, H3N 1X9\\
Montreal, QC Canada\\
\And
Soumyasundar Pal$^\ast$ \quad Mark Coates\\
Dept. Electrical and Computer Engineering\\
McGill University\\
3480 University St, H3A 0E9\\
Montreal, QC, Canada
\And
Deniz \"{U}stebay\\
Huawei Noah's Ark Lab\\
Montreal Research Centre\\
7101 Avenue du Parc, H3N 1X9\\
Montreal, QC Canada\\
}

\newcommand{\bx}{\bm{x}}
\newcommand{\by}{\bm{y}}

\newcommand{\mG}{\mathcal{G}}
\newcommand{\BX}{\mathbf{X}}
\newcommand{\BZ}{\mathbf{Z}}

\begin{document}

\maketitle

\begin{abstract}
  Recently, techniques for applying convolutional neural networks to
  graph-structured data have emerged. Graph convolutional neural
  networks (GCNNs) have been used to address node and graph
  classification and matrix completion. Although the performance has
  been impressive, the current implementations have limited capability
  to incorporate uncertainty in the graph structure. Almost all GCNNs
  process a graph as though it is a ground-truth depiction of the
  relationship between nodes, but often the graphs employed in
  applications are themselves derived from noisy data or modelling
  assumptions. Spurious edges may be included; other edges may be
  missing between nodes that have very strong relationships. In this
  paper we adopt a Bayesian approach, viewing the observed graph as a
  realization from a parametric family of random graphs. We then
  target inference of the joint posterior of the random graph
  parameters and the node (or graph) labels. We present the Bayesian
  GCNN framework and develop an iterative learning procedure for the
  case of assortative mixed-membership stochastic block models. We
  present the results of experiments that demonstrate that the
  Bayesian formulation can provide better performance when there are
  very few labels available during the training process.
\end{abstract}

\section{Introduction}

Novel approaches for applying convolutional neural networks
to graph-structured data have emerged in recent years. Commencing with
the work in~\cite{bruna2013,henaff2015}, there have been numerous
developments and improvements. Although these graph
convolutional neural networks (GCNNs) are promising, the current
implementations have limited capability to handle uncertainty in
the graph structure, and treat the graph topology as ground-truth
information. This in turn leads to an inability to adequately
characterize the uncertainty in the predictions made by the neural
network.

In contrast to this past work, we employ a Bayesian framework and view
the observed graph as a realization from a parametric random graph
family. The observed adjacency matrix is then used in conjunction with
features and labels to perform joint inference. The results reported
in this paper suggest that this formulation, although computationally
more demanding, can lead to an ability to learn more from less
data, a better capacity to represent uncertainty, and better
robustness and resilience to noise or adversarial attacks.

In this paper, we present the novel Bayesian GCNN framework and
discuss how inference can be performed. To provide a concrete example
of the approach, we focus on a specific random graph model, the
assortative mixed membership block model. We address the task of
semi-supervised classification of nodes and examine the resilience of
the derived architecture to random perturbations of the graph topology.

\section{Related work}
A significant body of research focuses on using neural networks to
analyze structured data when there is an underlying graph describing
the relationship between data items. Early work led to the development
of the {\em graph neural network
  (GNN)}~\cite{frasconi1998,scarselli2009,li2016b}. The GNN approaches
rely on recursive processing and propagation of information across
the graph. Training can often take a long time to converge and the
required time scales undesirably with respect to the
number of nodes in the graph, although recently an approach to
mitigate this has been proposed by~\cite{liao2018}. 

{\em Graph convolutional neural networks (GCNNs)} have emerged more
recently, with the first proposals in~\cite{bruna2013,henaff2015,duvenaud2015}. A spectral filtering approach was introduced
in~\cite{defferrard2016} and this method was simplified or improved
in~\cite{kipf2017,levie2017,chen2018}. Spatial filtering or aggregation
strategies were adopted in~\cite{atwood2016,hamilton2017b}. A general framework for training neural
networks on graphs and manifolds was presented by~\cite{monti2017} and
the authors explain how several of the other methods can be
interpreted as special cases.

The performance of the GCNNs can be improved by incorporating
attention nodes~\cite{velivckovic2018}, leading to the graph attention
network (GAT). Experiments have also demonstrated that gates, edge conditioning, and skip
connections can prove beneficial~\cite{bresson2017,sukhbaatar2016,simonovsky2017}.
In some problem settings it is also beneficial to consider an
ensemble of graphs~\cite{anirudh2017}, multiple adjacency matrices~\cite{such2017} or the dual
graph~\cite{monti2018}. Compared to this past work, the primary methodological
novelty in our proposed approach involves the adoption of a
Bayesian framework and the treatment of the observed graph as
additional data to be used during inference.

There is a rich literature on
Bayesian neural networks, commencing with pioneering
work~\cite{tishby1989,denker1991,mackay1992,neal1993} and extending to more
recent contributions~\cite{hernandez2015,gal2016,sun2017,louizos2017}. To
the best of our knowledge, Bayesian neural networks have not yet been developed
for the analysis of data on graphs. 

\section{Background}
\subsection{Graph convolutional neural networks (GCNNs)}
Although graph convolutional neural networks can be applied to a
variety of inference tasks, in order to make the description more
concrete we consider the task of identifying the labels of nodes in a
graph.  Suppose that we observe a graph
$\mathcal{G}_{obs} = (\mathcal{V},\mathcal{E})$, comprised of a set of
$N$ nodes $\mathcal{V}$ and a set of edges $\mathcal{E}$. For each
node we measure data (or derive features), denoted
$\bx_i$ for node $i$. For some subset of the nodes
$\mathcal{L}\subset \mathcal{V}$, we can also measure labels
$\mathbf{Y_{\mathcal{L}}} = \{\by_i: i \in \mathcal{L}\}$. In a classification context, the label
$\by_i$ identifies a category; in a regression context
$\by_i$ can be real-valued. Our task is to use the features $\bx$ and
the observed graph structure $\mathcal{G}_{obs}$ to estimate the labels of the unlabelled nodes.

A GCNN performs this task by performing graph convolution operations
within a neural network architecture. Collecting the feature vectors
as the rows of a matrix $\bm{X}$, the layers of a
GCNN~\cite{defferrard2016,kipf2017} are of the form:\

\begin{align}
\mathbf{H}^{(1)} &=\sigma(\mathbf{A}_{\mathcal{G}}\mathbf{X}\mathbf{W}^{(0)}) \\
\mathbf{H}^{(l+1)} &= \sigma(\mathbf{A}_{\mathcal{G}}\mathbf{H}^{(l)}\mathbf{W}^{(l)}) 
\end{align}
Here $\mathbf{W}^{(l)}$ are the weights of the neural network at layer
$l$, $\mathbf{H}^{(l)}$ are the output features from layer $l-1$, and
$\sigma$ is a non-linear activation function. The matrix
$\mathbf{A}_{\mathcal{G}}$ is derived from the observed graph and
determines how the output features are mixed across the graph at each
layer. The final output for an $L$-layer network is $\mathbf{Z} =
\mathbf{H}^{(L)}$. Training of the weights of the neural network is performed
by backpropagation with the goal of minimizing an error metric between
the observed labels $\mathbf{Y}$ and the network predictions $\mathbf{Z}$.
Performance improvements can be achieved by
enhancing the architecture with components that have proved
useful for standard CNNs, including attention
nodes~\cite{velivckovic2018}, and skip connections and
gates~\cite{li2016b,bresson2017}. 

Although there are many different flavours of GCNNs, all current
versions process the graph as though it is a ground-truth
depiction of the relationship between nodes. This is
despite the fact that in many cases the graphs employed
in applications are themselves derived from noisy data or
modelling assumptions. Spurious edges may be included;
other edges may be missing between nodes that have very
strong relationships. Incorporating attention mechanisms as in~\cite{velivckovic2018}
addresses this to some extent; attention nodes can learn that some
edges are not representative of a meaningful relationship and reduce
the impact that the nodes have on one another. But the attention
mechanisms, for computational expediency, are limited to processing
existing edges --- they cannot create an edge where one should
probably exist. This is also a limitation of the ensemble approach
of~\cite{anirudh2017}, where learning is performed on multiple graphs
derived by erasing some edges in the graph.

\subsection{Bayesian neural networks}
\label{subsec:bayes_nn}
We consider the case where we have training inputs
$\BX=\{x_1 , . . . , x_n\}$ and corresponding outputs
$\mathbf{Y} = \{y_1 , . . . , y_n\}$. Our goal is to learn a function
$y = f (x)$ via a neural network with fixed configuration (number of
layers, activation function, etc., so that the weights are sufficient
statistics for $f$) that provides a likely explanation for the
relationship between $x$ and $y$. The weights $W$ are modelled as
random variables in a Bayesian approach and we introduce a prior
distribution over them. Since $W$ is not deterministic, the output of
the neural network is also a random variable. Prediction for a new
input $x$ can be formed by integrating with respect to the posterior
distribution of $W$ as follows:
\begin{align}
p(y|x,\BX,\mathbf{Y}) = \int p(y|x,W)p(W|\BX,\mathbf{Y})\,dW\label{eq:BayesianNN}\,.
\end{align}
The term $p(y|x,W)$ can be viewed as a likelihood; in a classification
task it is modelled using a categorical distribution by applying a
softmax function to the output of the neural network; in a regression
task a Gaussian likelihood is often an appropriate choice.  The
integral in eq. \eqref{eq:BayesianNN} is in general
intractable. Various techniques for inference of $p(W|\BX,\mathbf{Y})$
have been proposed in the literature, including expectation propagation
\cite{hernandez2015}, variational inference
\cite{gal2016,sun2017,louizos2017}, and Markov Chain Monte Carlo
methods \cite{neal1993,korattikara2015,li2016d}. In particular, in
\cite{gal2016}, it was shown that with suitable variational
approximation for the posterior of $W$, Monte Carlo dropout is
equivalent to drawing samples of $W$ from the approximate posterior
and eq. \eqref{eq:BayesianNN} can be approximated by a Monte Carlo
integral as follows:
\begin{align}
p(y|x,\BX,\mathbf{Y}) \approx \dfrac{1}{T} \displaystyle \sum_{i=1}^{S} p(y|x,W_i)\,,
\end{align}
where $S$ weights $W_i$ are obtained via dropout. 

\section{Methodology}

We consider a Bayesian approach, viewing the observed graph as a
realization from a parametric family of random graphs. We then target
inference of the joint posterior of the random graph parameters,
weights in the GCNN and the node (or graph) labels. Since we are
usually not directly interested in inferring the graph parameters,
posterior estimates of the labels are obtained by marginalization. The
goal is to compute the posterior probability of labels, which can be
written as:
\begin{align}
p(\BZ|\mathbf{Y_{\mathcal{L}}},\BX,\mG_{obs}) &= \int p(\BZ|W,\mG,\BX) p(W|\mathbf{Y_{\mathcal{L}}},\BX,\mG)\,\nonumber\\
& \quad \quad p(\mG|\lambda) p(\lambda|\mG_{obs}) \,dW\,d\mG\,d\lambda\label{eq:exact_posterior}\,.
\end{align}
Here $W$ is a random variable representing the weights of a Bayesian
GCNN over graph $\mG$, and $\lambda$ denotes the parameters that
characterize a family of random graphs. The term $p(\BZ|W,\mG,\BX)$
can be modelled using a categorical distribution by applying a softmax
function to the output of the GCNN, as discussed above.

This integral in eq.~\eqref{eq:exact_posterior} is intractable. We can
adopt a number of strategies to approximate it, including variational
methods and Markov Chain Monte Carlo (MCMC). For example, in order to
approximate the posterior of weights
$p(W|\mathbf{Y_{\mathcal{L}}},\BX,\mG)$, we could use variational
inference \cite{gal2016,sun2017,louizos2017} or MCMC \cite{neal1993,korattikara2015,li2016d}. Various parametric random graph generation
models can be used to model $p(\lambda|\mG_{obs})$, for example a
stochastic block model \cite{peixoto2017}, a mixed membership
stochastic block model \cite{airoldi2009}, or a degree corrected block
model \cite{peng2016}. For inference of $p(\lambda|\mG_{obs})$, we can
use MCMC \cite{li2016c} or variational inference \cite{gopalan2012}.

 A Monte Carlo approximation of eq. \eqref{eq:exact_posterior} is:
\begin{align}
  &p(\BZ|\mathbf{Y_{\mathcal{L}}},\BX,\mG_{obs}) \approx \nonumber \\
&\quad\quad\quad  \dfrac{1}{V} \sum_v^V \dfrac{1}{N_G S}\sum_{i=1}^{N_G}\sum_{s=1}^S p(\BZ|W_{s,i,v},\mG_{i,v},\BX) \,.
\label{eq:MC_posterior}
\end{align}
In this approximation, $V$ samples $\lambda_v$ are drawn from
$p(\lambda|\mG_{obs})$; the precise method for generating these
samples from the posterior varies depending on the nature of the graph
model. The $N_G$ graphs $\mG_{i,v}$ are sampled from $p(\mG|\lambda_v)$ using the adopted random
graph model. $S$ weight matrices $W_{s,i,v}$ are sampled
from $p(W|\mathbf{Y_{\mathcal{L}}},\BX,\mG_{i,v})$ from the Bayesian
GCN corresponding to the graph $\mG_{i,v}$. 

\subsection{Example: Assortative mixed membership stochastic block model}

For the Bayesian GCNNs derived in this paper, we use an assortative
mixed membership stochastic block model (a-MMSBM) for the graph
\cite{gopalan2012,li2016c} and learn its parameters $\lambda = \{\pi,\beta\}$
using a stochastic optimization approach. The assortative MMSBM,
described in the following section, is a good choice to model a graph
that has relatively strong community structure (such as the citation
networks we study in the experiments section). It generalizes the
stochastic block model by allowing nodes to belong to more than one
community and to exhibit assortative behaviour, in the sense that a
node can be connected to one neighbour because of a relationship
through community A and to another neighbour because of a relationship
through community B.

Since $\mG_{obs}$ is often noisy and may not fit the adopted parametric block
model well, sampling $\pi_v$ and $\beta_v$ from
$p(\pi,\beta|\mG_{obs})$ can lead to high variance. This can lead to
the sampled graphs $\mG_{i,v}$ being very different from
$\mG_{obs}$. Instead, we replace the integration over $\pi$ and $\beta$
with a maximum a posteriori estimate \cite{mackay1996}. We approximately compute
\begin{align}
     \{\hat{\pi},\hat{\beta}\} &= \arg\max_{\beta,\pi}  p(\beta,\pi|\mG_{obs})
 \end{align}
 by incorporating suitable priors over $\beta$ and $\pi$ and use the approximation:
\begin{align}
p(\BZ|\mathbf{Y_{\mathcal{L}}},\BX,\mG_{obs}) \approx \dfrac{1}{N_G S}\sum_{i=1}^{N_G}\sum_{s=1}^S p(\BZ|W_{s,i},\mG_{i},\BX)\,.\label{eq:map_approx}
\end{align}
In this approximation, $W_{s,i}$ are approximately sampled from $p(W|\mathbf{Y_{\mathcal{L}}},\BX,\mG_{i})$ using Monte Carlo dropout over the Bayesian GCNN corresponding to $\mG_{i}$. The $\mG_{i}$ are sampled from $p(\mG|\hat{\pi},\hat{\beta})$.

\subsection{Posterior inference for the MMSBM}
\label{subsec:mmsbm}
For the undirected observed graph $\mG_{obs} = \{y_{ab} \in \{0,1\}: 1 \leq a <b \leq N\}$, $y_{ab} = 0$ or $1$ indicates absence or presence of a link between node $a$ and node $b$. In MMSBM, each node $a$ has a $K$ dimensional community membership probability distribution $\pi_a = [\pi_{a1},...\pi_{aK}]^T$, where $K$ is the number of categories/communities of the nodes. For any two nodes $a$ and $b$, if both of them belong to the same community, then the probability of a link between them is significantly higher than the case where the two nodes belong to different communities \cite{airoldi2009}. The generative model is described as:\\

For any two nodes $a$ and $b$,
\begin{itemize}
\item Sample $z_{ab} \sim \pi_a$ and $z_{ba} \sim \pi_b$.
\item If $z_{ab} = z_{ba} = k$, sample a link $y_{ab} \sim \text{Bernoulli}(\beta_k)$. Otherwise, $y_{ab} \sim \text{Bernoulli}(\delta)$.
\end{itemize}

Here, $0 \leq \beta_k \leq 1$ is termed community strength of the $k$-th community and $\delta$ is the cross community link probability, usually set to a small value. The joint posterior of the parameters $\pi$ and $\beta$ is given as:

\begin{align}
&p(\pi,\beta|\mG_{obs}) \propto p(\beta)p(\pi)p(\mG_{obs}|\pi, \beta)\,\nonumber\\
=&\prod_{k=1}^{K} p(\beta_k) \prod_{a=1}^{N} p(\pi_a) \prod_{1 \leq a<b \leq N}\sum_{z_{ab},z_{ba}} p(y_{ab},z_{ab},z_{ba}|\pi_a,\pi_b,\beta)\,.\label{eq:mmsbm_posterior}
\end{align}

We use a $\text{Beta}(\eta)$ distribution for the prior of $\beta_k$ and a
Dirichlet distribution, $\text{Dir}(\alpha)$, for the prior of $\pi_a$, where $\eta$
and $\alpha$ are hyper-parameters.

\subsection{Expanded mean parameterisation}

Maximizing the posterior of eq. \eqref{eq:mmsbm_posterior} is a
constrained optimization problem with $\beta_k, \pi_{ak} \in (0,1)$
and $\displaystyle \sum_{k=1}^{K}\pi_{ak} =1$. Employing a standard iterative algorithm
with a gradient based update rule does not guarantee that the
constraints will be satisfied. Hence we consider an expanded mean
parameterisation \cite{patterson2013} as follows. We introduce the
alternative parameters $\theta_{k0}, \theta_{k1} \geq 0$ and adopt as
the prior for these parameters a product of independent
$\text{Gamma}(\eta ,\rho)$ distributions. These parameters are related
to the original parameter $\beta_k$ through the relationship
$\beta_k = \dfrac{\theta_{k1}}{\theta_{k0} +\theta_{k1}}$. This results
in a $\text{Beta}(\eta)$ prior for $\beta_k$. Similarly, we
introduce a new parameter $\phi_a \in \mathbb{R}^K_{+}$ and adopt as
its prior a product of independent
$\text{Gamma}(\alpha,\rho)$ distributions. We define
$\pi_{ak} = \dfrac{\phi_{ak}}{\displaystyle \sum_{l=1}^{K}
  \phi_{al}}$, which results in a Dirichlet prior, $\text{Dir}(\alpha)$, for
$\pi_a$. The boundary conditions $\theta_{ki}, \phi_{ak} \geq 0$ can
be handled by simply taking the absolute value of the update.

\subsection{Stochastic optimization and mini-batch sampling}

We use preconditioned gradient ascent to maximize the joint posterior
in eq. \eqref{eq:mmsbm_posterior} over $\theta$ and $\phi$. In many
graphs that are appropriately modelled by a stochastic block model, 
most of the nodes belong strongly to only one of the $K$ communities,
so the MAP estimate for many $\pi_a$ lies near one of the corners of
the probability simplex. This suggests that the scaling of different
dimensions of $\phi_a$ can be very different. Similarly, as
$\mG_{obs}$ is typically sparse, the community strengths $\beta_k$ are
very low, indicating that the scales of $\theta_{k0}$ and $\theta_{k1}$ are
very different. We use preconditioning matrices
$G(\theta) = \mathrm{diag}(\theta)^{-1}$ and $G(\phi) = \mathrm{diag}(\phi)^{-1}$ as in
\cite{patterson2013}, to obtain the following update rules:
\begin{align}
  \theta_{ki}^{(t+1)} &= \nonumber\\
  &\Big|\theta_{ki}^{(t)} + \epsilon_t\Big(\eta - 1 -\rho \theta_{ki}^{(t)} +  \theta_{ki}^{(t)}\sum_{a=1}^{N}\sum_{b= a+1}^{N} g_{ab}(\theta_{ki}^{(t)}) \Big) \Big|\label{eq:theta_update}\,,\\
    \phi_{ak}^{(t+1)} &= \Big|\phi_{ak}^{(t)} + \epsilon_t\Big(\alpha -1 -\rho \phi_{ak}^{(t)}\sum_{b=1,b \neq a}^{N} g_{ab}(\phi_{ak}^{(t)}) \Big) \Big|\label{eq:phi_update}\,,
\end{align}
where $\epsilon_t = \epsilon_0(t+\tau)^{-\kappa}$ is a decreasing
step-size, and $g_{ab}(\theta_{ki})$ and $g_{ab}(\phi_{ak})$ are the
partial derivatives of $\log p(y_{ab}|\pi_a,\pi_b,\beta)$
w.r.t. $\theta_{ki}$ and $\phi_{ak}$, respectively. Detailed
expressions for these derivatives are provided in eqs. (9) and (14) of \cite{li2016c}.

Implementation of \eqref{eq:theta_update} and
\eqref{eq:phi_update} is $\mathcal{O}(N^2K)$ per iteration, where $N$
is the number of nodes in the graph and $K$ the number of
communities. This can be prohibitively expensive for large graphs. We
instead employ a stochastic gradient based strategy as follows. For
update of $\theta_{ki}$'s in eq.~\eqref{eq:theta_update}, we split the
$\mathcal{O}(N^2)$ sum over all edges and non-edges,
$\sum_{a=1}^{N}\sum_{b= a+1}^{N}$, into two separate terms. One of
these is a sum over all observed edges and the other is a sum over all
non-edges. We calculate the term corresponding to observed edges
exactly (in the sparse graphs of interest, the number of edges is
closer to $\mathcal{O}(N)$ than $\mathcal{O}(N^2)$). For the other
term we consider a mini-batch of 1 percent of randomly sampled
non-edges and scale the sum by a factor of 100.

At any single iteration, we update the $\phi_{ak}$ values for only $n$ randomly
sampled nodes ($n<N$), keeping the rest of them fixed. For the update of
$\phi_{ak}$ values of any of the randomly selected $n$ nodes, we split the
sum in eq.~\eqref{eq:phi_update} into two terms. One involves all of the
neighbours (the set of neighbours of node $a$ is denoted by
$\mathcal{N}(a)$) and the other involves all the non-neighbours of
node $a$. We calculate the first term exactly. For the second
term, we use $n-|\mathcal{N}(a)|$ randomly sampled
non-neighbours and scale the sum by a factor of
$\dfrac{N-1-|\mathcal{N}(a)|}{n-|\mathcal{N}(a)|}$ to maintain
unbiasedness of the stochastic gradient. Overall the update of the $\phi$
values involve $\mathcal{O}(n^2K)$ operations instead of
$\mathcal{O}(N^2K)$ complexity for a full batch update.

Since the posterior in the MMSBM is very high-dimensional, random
initialization often does not work well. We train a GCNN~\cite{kipf2017} on $\mG_{obs}$ and use its softmax output to
initialize $\pi$ and then initialize $\beta$ based on the block
structure imposed by $\pi$. The resulting algorithm is given in
Algorithm \ref{alg:Bayesian-GCNN}.

\begin{algorithm}[H]
 \hspace*{\algorithmicindent} \textbf{Input: $\mathcal{G}_{obs}$, $\BX$, $\mathbf{Y_{\mathcal{L}}}$} \\
 \hspace*{\algorithmicindent} \textbf{Output: $p(\BZ|\mathbf{Y_{\mathcal{L}}},\BX,\mG_{obs})$}
\begin{algorithmic}[1]
\STATE Initialization: train a GCNN to initialize the inference in
MMSBM and the weights in the Bayesian GCNN.

\FOR{$i = 1:N_G$}

\STATE Perform $N_b$ iterations of MMSBM inference to obtain $(\hat{\pi},\hat{\beta})$.

\STATE Sample graph $\mG_i \sim p(\mG|\hat{\pi},\hat{\beta})$.

\FOR{$s=1:S$}

\STATE Sample weights $W_{s,i}$ via MC dropout by training a GCNN over the graph $\mG_i$.

\ENDFOR

\ENDFOR

\STATE Approximate $p(\BZ|\mathbf{Y_{\mathcal{L}}},\BX,\mG_{obs})$ using eq. \eqref{eq:map_approx}.
\end{algorithmic}
\caption{Bayesian-GCNN}
\label{alg:Bayesian-GCNN}
\end{algorithm}

\section{Experimental Results}
We explore the performance of the proposed Bayesian GCNN on three well-known citation datasets~\cite{sen2008}: Cora, CiteSeer, and Pubmed. In these datasets each node represents a document and has a
sparse bag-of-words feature vector associated with it. Edges are
formed whenever one document cites another. The direction of
the citation is ignored and an undirected graph with a symmetric
adjacency matrix is constructed. Each node label represents
the topic that is associated with the document. We assume that we have access to several labels per
class and the goal is to predict the unknown document labels. The statistics of these datasets are represented in Table~\ref{table:datasets}.

\begin{table}[h]
	\centering
	\footnotesize{
		\begin{tabular}{lccc}			
			\toprule[0.25ex]
			\textbf{ } &\textbf{Cora} & \textbf{CiteSeer} & \textbf{Pubmed} \\ 
			\midrule
			\textbf{Nodes} & 2708 & 3327 & 19717 \\
			\textbf{Edges} & 5429 & 4732 & 44338 \\
			\textbf{Features per node} & 1433 & 3703 & 500 \\
			\textbf{Classes} & 7 & 6 & 3 \\
			\bottomrule[0.25ex]
		\end{tabular}
	}
	\caption{Summary of the datasets used in the experiments.}
	\label{table:datasets}
\end{table}

The hyperparameters of GCNN are the same for all of the experiments and are
based on~\cite{kipf2017}. The GCNN has two layers where the number of
hidden units is 16, the learning rate is 0.01, the L2 regularization parameter
is 0.0005, and the dropout rate is 50\% at each layer. These
hyperparameters are also used in the Bayesian GCNN. In addition, the
hyperparameters associated with MMSBM inference are set as follows: $\eta=1, \alpha=1,\rho =0.001, n = 500, \epsilon_0 = 1, \tau = 1024$
and $\kappa=0.5$.

\subsection{Semi-supervised node classification}

We first evaluate the performance of the proposed Bayesian GCNN
algorithm and compare it to the state-of-the-art methods on the
semi-supervised node classification problem. In addition to the 20
labels per class training setting explored in previous
work~\cite{kipf2017,velivckovic2018}, we also evaluate the performance
of these algorithms under more severely limited data scenarios where
only 10 or 5 labels per class are available.

The data is split into train and test datasets in two different
ways. The first is the fixed data split originating
from~\cite{yang2016}. In 5 and 10 training labels per class cases, we
construct the fixed split of the data by using the first 5 and 10 labels in the
original partition of~\cite{yang2016}. The second type of split is
random where the training and test sets are created at random for each
run. This provides a more robust comparison of the
model performance as the specific split of data can have a significant impact in
the limited training labels case.


We compare ChebyNet~\cite{defferrard2016}, GCNN~\cite{kipf2017}, and
GAT~\cite{velivckovic2018} to the Bayesian GCNN proposed
in this paper. Tables~\ref{table:cora},~\ref{table:citeseer},~\ref{table:pubmed}
show the summary of results on Cora, Citeseer and Pubmed datasets
respectively.  The results are from 50 runs with random weight
initializations. The standard errors in the fixed split scenarios are due
to the random initialization of weights whereas the random split
scenarios have higher variance due to the additional randomness induced by
the split of data. We conducted Wilcoxon signed rank  tests to
evaluate the significance of the difference between the
best-performing algorithm and the second-best. The asterisks in the
table indicate the scenarios where the score differentials were
statistically significant for a p-value threshold of 0.05. 

\begin{table}[h]
	\centering
	\footnotesize{
		\begin{tabular}{lcccc}					
			\toprule[0.25ex]
			\textbf{Random split} &\textbf{5 labels}        & \textbf{10 labels}         & \textbf{20 labels} \\ 
			\midrule
			\textbf{ChebyNet}             &61.7$\pm$6.8            &72.5$\pm$3.4              &78.8$\pm$1.6        \\
			\textbf{GCNN}             &70.0$\pm$3.7            & 76.0$\pm$2.2            &79.8$\pm$1.8  \\
			\textbf{GAT}              &70.4$\pm$3.7            &76.6$\pm$2.8              &79.9$\pm$1.8   \\
			\textbf{Bayesian GCN}     &\!\!$\bm{^\ast}$\textbf{74.6$\pm$2.8}   &\!\!$\bm{^\ast}$\textbf{77.5$\pm$2.6}     &\textbf{80.2$\pm$1.5} \\
			\midrule
			\textbf{Fixed split}  &\textbf{}                 & \textbf{}                  & \textbf{}\\
			\midrule
			\textbf{ChebyNet}             &67.9$\pm$3.1            &72.7$\pm$2.4             &80.4$\pm$0.7        \\
			\textbf{GCNN}             &74.4$\pm$0.8            &74.9$\pm$0.7              &\textbf{81.6$\pm$0.5}       \\
			\textbf{GAT}              &73.5$\pm$2.2            &74.5$\pm$1.3              &81.6$\pm$0.9  \\
			\textbf{Bayesian GCN}     &\!\!$\bm{^\ast}$\textbf{75.3$\pm$0.8}   &\!\!$\bm{^\ast}$\textbf{76.6$\pm$0.8}    &81.2$\pm$0.8     \\
			\bottomrule[0.25ex]
		\end{tabular}
	}
	\caption{Prediction accuracy (percentage of correctly
          predicted labels) for Cora dataset. Asterisks denote
          scenarios where a Wilcoxon signed rank test indicates a
        statistically significant difference between the scores of the
      best and second-best algorithms.}
	\label{table:cora}
\end{table}

\begin{table}[h]
	\centering
	\footnotesize{
		\begin{tabular}{lcccc}
			\toprule[0.25ex]
			\textbf{Random split}  &\textbf{5 labels}        & \textbf{10 labels}         & \textbf{20 labels} \\ 
			\midrule
			\textbf{ChebyNet}             &58.5$\pm$4.8            &65.8$\pm$2.8              &67.5$\pm$1.9             \\
			\textbf{GCNN}             &58.5$\pm$4.7            &65.4$\pm$2.6              &67.8$\pm$2.3        \\
			\textbf{GAT}              &56.7$\pm$5.1            &64.1$\pm$3.3              &67.6$\pm$2.3       \\
			\textbf{Bayesian GCN}     &\!\!$\bm{^\ast}$\textbf{63.0$\pm$4.8}   &\!\!$\bm{^\ast}$\textbf{69.9$\pm$2.3}     &\!\!$\bm{^\ast}$\textbf{71.1$\pm$1.8}   \\
			\midrule
			\textbf{Fixed split}  &\textbf{}                 & \textbf{}                  & \textbf{}\\
			\midrule
			\textbf{ChebyNet}             &53.0$\pm$1.9            &67.7$\pm$1.2              &70.2$\pm$0.9        \\
			\textbf{GCNN}             &55.4$\pm$1.1            &65.8$\pm$1.1              &70.8$\pm$0.7       \\
			\textbf{GAT}              &55.4$\pm$2.6            &66.1$\pm$1.7              &70.8$\pm$1.0    \\
			\textbf{Bayesian GCN}     &\!\!$\bm{^\ast}$\textbf{57.3$\pm$0.8}   &\!\!$\bm{^\ast}$\textbf{70.8$\pm$0.6}     &\!\!$\bm{^\ast}$\textbf{72.2$\pm$0.6}\\
			\bottomrule[0.25ex]
		\end{tabular}
	}
	\caption{Prediction accuracy (percentage of correctly
          predicted labels) for Citeseer dataset. Asterisks denote
          scenarios where a Wilcoxon signed rank test indicates a
        statistically significant difference between the scores of the
      best and second-best algorithms.}
	\label{table:citeseer}
\end{table}

\begin{table}[h]
	\centering
	\footnotesize{
		\begin{tabular}{lcccc}
			\toprule[0.25ex]
			\textbf{Random split}  &\textbf{5 labels}        &\textbf{10 labels}         &\textbf{20 labels} \\ 
			\midrule
			\textbf{ChebyNet}             &62.7$\pm$6.9            &68.6$\pm$5.0              &74.3$\pm$3.0             \\
			\textbf{GCNN}              &69.7$\pm$4.5           &\!\!$\bm{^\ast}$\textbf{73.9$\pm$3.4}    &\!\!$\bm{^\ast}$\textbf{77.5$\pm$2.5}  \\
			\textbf{GAT}               &68.0$\pm$4.8           &72.6$\pm$3.6             &76.4$\pm$3.0\\
			\textbf{Bayesian GCNN}     &\textbf{70.2$\pm$4.5}  &73.3$\pm$3.1             &{76.0$\pm$2.6}  \\
			\midrule
			\textbf{Fixed split}  &\textbf{}                 & \textbf{}                 &\textbf{}\\
			\midrule
			\textbf{ChebyNet}             &68.1$\pm$2.5            &69.4$\pm$1.6              &76.0$\pm$1.2             \\
			\textbf{GCNN}             &69.7$\pm$0.5            &\!\!$\bm{^\ast}$\textbf{72.8$\pm$0.5}            &\!\!$\bm{^\ast}$\textbf{78.9$\pm$0.3}  \\
			\textbf{GAT}              &70.0$\pm$0.6   & 71.6$\pm$0.9            &76.9$\pm$0.5 \\
			\textbf{Bayesian GCNN}    &\!\!$\bm{^\ast}$\textbf{70.9$\pm$0.8}           &72.3$\pm$0.8    &76.6$\pm$0.7\\
			\bottomrule[0.25ex]
		\end{tabular}
	}
	\caption{Prediction accuracy (percentage of correctly predicted labels) for Pubmed dataset. Asterisks denote
          scenarios where a Wilcoxon signed rank test indicates a
        statistically significant difference between the scores of the
      best and second-best algorithms.}
	\label{table:pubmed}
\end{table}

Note that the implementation of the GAT method as provided by the
authors employs a validation set of 500 examples which is used to
monitor validation accuracy. The model that yields the minimum
validation error is selected as final model. We report results
without this validation set monitoring as large validation sets are
not always available and the other methods examined here do not
require one.

The results of our experiments illustrate the improvement in
classification accuracy provided by Bayesian GCNN for Cora and
Citeseer datasets in the random split scenarios. The improvement is
more pronounced when the number of available labels is limited to 10
or 5. In addition to increased accuracy, Bayesian GCNN provides lower
variance results in most tested scenarios. For the Pubmed dataset, the
Bayesian GCNN provides the best performance for the 5-label case, but
is outperformed by other techniques for the 10- and 20-label
cases. The Pubmed dataset has a much lower intra-community density
than the other datasets and a heavy-tailed degree distribution. The
assortative MMSBM is thus a relatively poor choice for the observed
graph, and this prevents the Bayesian approach from improving
the prediction accuracy.

In order to provide some insight into the information available from the
posterior of the MMSBM, we examined the 50 observed edges with lowest
average posterior probability for both the Cora and Citeseer
graphs. In the majority of cases the identified edges were inter-community
(connecting edges with different labels) or had one node with very low
degree (lower than 2). This accounted for 39 of the 50 edges for Cora
and 42 of the 50 edges for Citeseer. For the unobserved edges, we
analyzed the most probable edges from the posterior. Most of these are
intra-community edges (connecting nodes with the same label). For Cora
177 of the 200 edges identified as most probable are intra-community,
and for Citeseer 197 of 200.

\subsection{Classification under node attacks}
Several studies have shown the vulnerability of deep neural networks
to adversarial examples~\cite{goodfellow2015}. For graph convolutional
neural networks,~\cite{zugner2018} recently introduced a method to
create adversarial attacks that involve limited perturbation of the
input graph. The aim of the study was to demonstrate the vulnerability
of the graph-based learning algorithms.  Motivated by this study we
use a random attack mechanism to compare the robustness of GCNN and
Bayesian GCNN algorithms in the presence of noisy edges.

\textbf{Random node attack mechanism}: In each experiment, we target
one node to attack. We choose a fixed number of perturbations
$\Delta = d_{v_0} + 2$, where $v_0$ is the node we want to attack, and
$d_{v0}$ is the degree of this target node. The random attack involves
removing $(d_{v_0} + 2)/2$ nodes from the target node's set of
neighbors, and sampling $(d_{v_0} + 2)/2$ cross-community edges
(randomly adding neighbors that have different labels than the target
node) to the target node. For each target node, this procedure is
repeated five times so that five perturbed graphs are generated. There
are two types of adversarial mechanisms in~\cite{zugner2018}. In the
first type, called an evasion attack, data is modified to fool an
already trained classifier, and in the second, called a poisoning
attack, the perturbation occurs before the model training. All of our
experiments are performed in the poisoning attack fashion.

\textbf{Selection of target node}: Similar to the setup
in~\cite{zugner2018}, we choose 40 nodes from the test set that are
correctly classified and simulate attacks on these nodes. The margin of classification for node $v$ is defined as:
\begin{align}
        \text{margin}_{v} = \text{score}_v(c_{true}) - \underset{c \neq c_{true}}{\max} \text{score}_v(c)\,,\nonumber
\end{align}
where $c_{true}$ is the true class of node $v$ and $\mathrm{score}_v$
denotes the classification score vector reported by the classifier for
node $v$. A correct classification leads to a positive margin; incorrect
classifications are associated with negative margins.
For each algorithm we choose the 10 nodes with the highest margin of
classification and 10 nodes with the lowest positive margin of
classification. The remaining 20 nodes are selected at random from the
set of nodes correctly classified by both algorithms. Thus, among the
40 target nodes, the two algorithms are sharing at least 20 common
nodes.

\textbf{Evaluation}: For each targeted node, we run the algorithm for
5 trials. The results of this experiment are summarized in
Tables~\ref{table:cora_attack} and~\ref{table:citeseer_attack}. These
results illustrate average performance over the target nodes and the
trials. Note that the accuracy figures in these tables are different
from Table~\ref{table:cora} and~\ref{table:citeseer} as here we are
reporting the accuracy for the 40 selected target nodes instead of the
entire test set.

\vspace{0.2cm}

\begin{table}[h]
	\centering
	\footnotesize{
		\begin{tabular}{ccc}
			\toprule[0.25ex]
			\textbf{ }                               &\textbf{No attack} & \textbf{Random attack} \\ 
			\midrule
			&\multicolumn{2}{c}{\textbf{Accuracy}} \\
			\midrule
			\textbf{GCNN}                            & 85.55\%           & 55.50\%    \\
			\textbf{Bayesian GCNN}                   & 86.50\%           & 69.50\%    \\
			\midrule
			&\multicolumn{2}{c}{\textbf{Classifier margin}  } \\
			\midrule
			\textbf{GCNN}                            & 0.557             & 0.152  \\
			\textbf{Bayesian GCNN}                   & 0.616            & 0.387      \\
			\bottomrule[0.25ex]
		\end{tabular}
	}
	\caption{Comparison of accuracy and classifier margins for the no attack and random attack scenarios on the Cora dataset. The results are for 40 selected target nodes and 5 runs of the algorithms for each target.}
	\label{table:cora_attack}
\end{table}

\vspace{0.2cm}

\begin{table}[h]
	\centering
	\footnotesize{
		\begin{tabular}{ccc}
			\toprule[0.25ex]
			\textbf{ }                               &\textbf{No attack} & \textbf{Random attack} \\ 
			\midrule
			&\multicolumn{2}{c}{\textbf{Accuracy}} \\
			\midrule
			\textbf{GCNN}                            & 88.5\%           & 43.0\%    \\
			\textbf{Bayesian GCNN}                   & 87.0\%           & 66.5\%    \\
			\midrule
			&\multicolumn{2}{c}{\textbf{Classifier margin}  } \\
			\midrule
			\textbf{GCNN}                            & 0.448             & 0.014  \\
			\textbf{Bayesian GCNN}                   & 0.507           & 0.335      \\
			\bottomrule
		\end{tabular}
	}
	\caption{Comparison of accuracy and classifier margins for the no attack and random attack scenarios on the Citeseer dataset. The results are for 40 selected target nodes and 5 runs of the algorithms for each target.}
	\label{table:citeseer_attack}
\end{table}

Overall the attacks affect both algorithms severely. GCNN loses 30\%
in prediction accuracy for the Cora dataset and 44.5\% for Citeseer
whereas the drop in prediction accuracy is more limited for Bayesian
GCNN with 17\% for Cora and 20.5\% for the Citeseer dataset. The
Bayesian GCNN is able to maintain the classifier margin much better 
compared to GCNN. For the Citeseer dataset the random attacks
almost eliminate the GCNN margin whereas Bayesian GCNN suffers a 34\%
decrease, but retains a positive margin on average.
 
Figure 1 provides further insight concerning the impact of the attack
on the two algorithms. The figure depicts the distribution of average
classifier margins over the targeted nodes before and after the random
attacks. Each circle in the figure shows the margin for one target
node averaged over the 5 random perturbations of the graph. Note that
some of the nodes have a negative margin prior to the random attack
because we select the correctly classified nodes with lowest average
margin based on 10 random trials and then perform another 5 random
trials to generate the depicted graph. We see that for GCNN the
attacks cause nearly half of the target nodes to be wrongly classified
whereas there are considerably fewer prediction changes for the
Bayesian GCNN. 

\begin{figure}[h]
	\centering
	 \begin{subfigure}[b]{0.46\textwidth}
	 	\centering
	 	\includegraphics[width=1\linewidth]{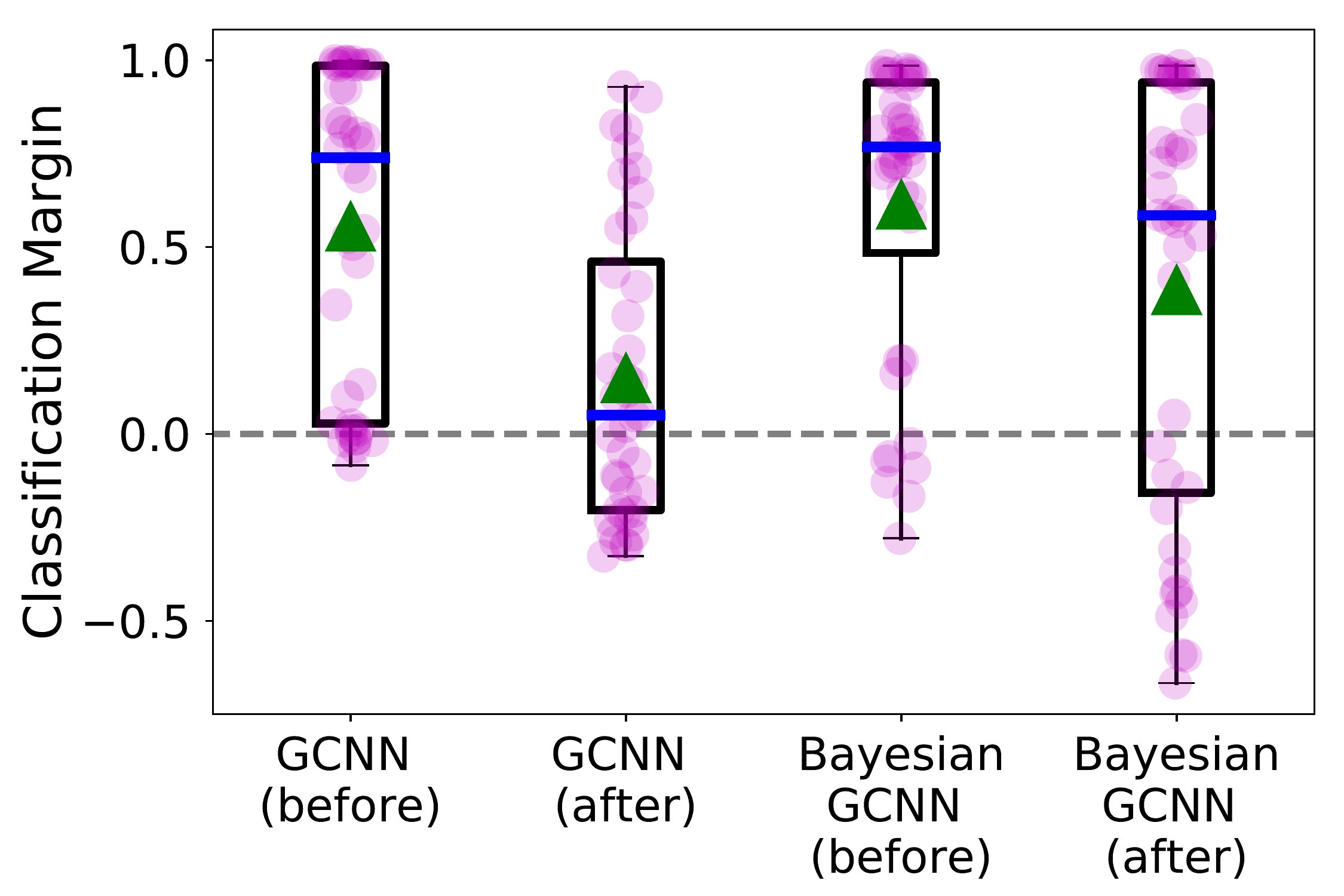}\caption{}
\end{subfigure}
	 	
	 	\begin{subfigure}[b]{0.46\textwidth}	
	 		\centering
	\includegraphics[width=1\linewidth]{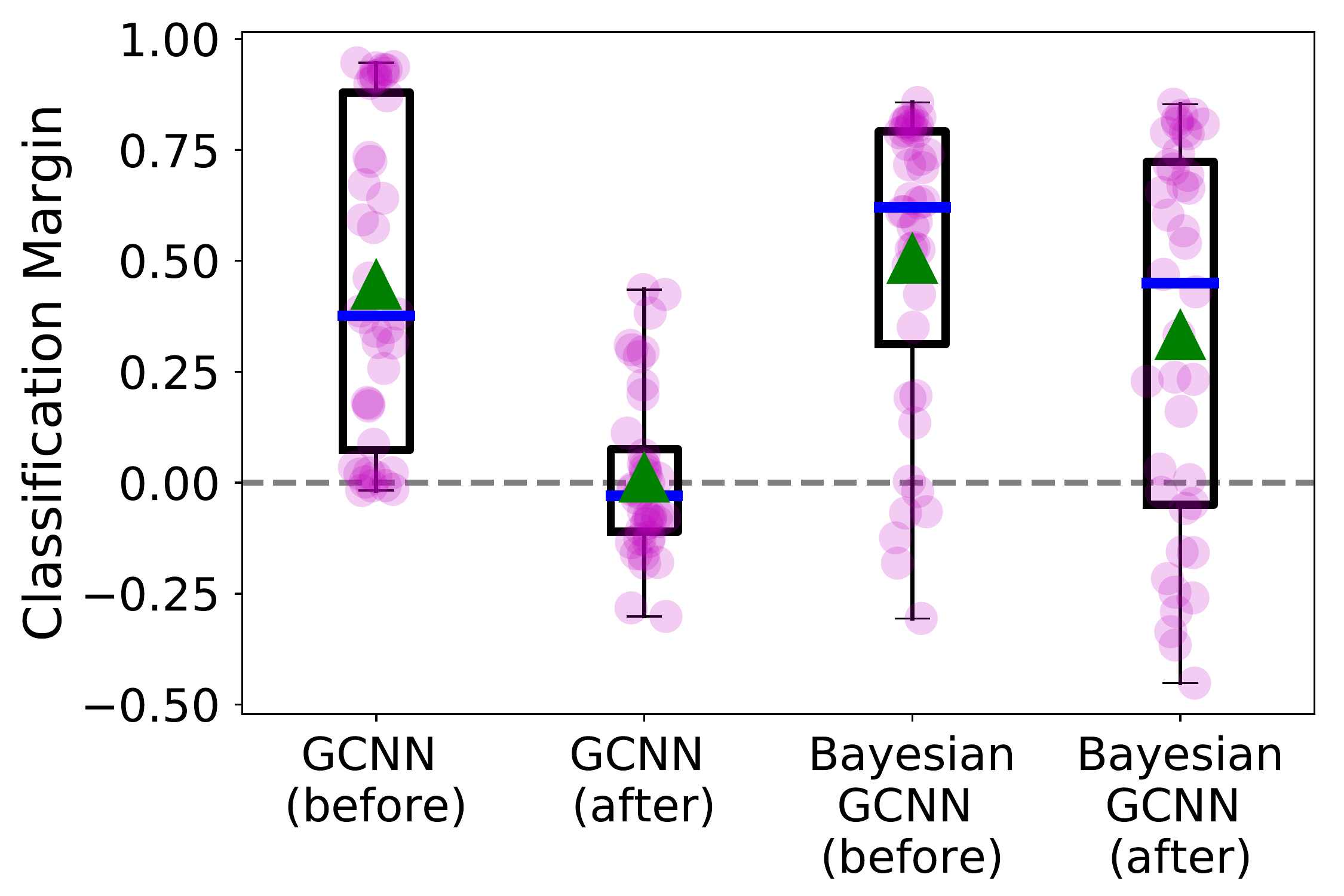}\caption{}
\end{subfigure}
	\caption{Boxplots of the average classification margin for 40 nodes before and after random attacks for GCNN and Bayesian GCNN on (a) Cora dataset  and (b) Citeseer dataset. The box indicates 25-75 percentiles; the triangle represents the mean value; and the median is shown by a horizontal line. Whiskers extend to the minimum and maximum of data points.}
	\label{fig:attack}
\end{figure}

\section{Conclusions and Future Work}

In this paper we have presented Bayesian graph convolutional neural
networks, which provide an approach for incorporating uncertain
graph information through a parametric random graph model. We 
provided an example of the framework for the case of an assortative
mixed membership stochastic block model and explained how approximate
inference can be performed using a combination of stochastic
optimization (to obtain maximum a posteriori estimates of the random
graph parameters) and approximate variational inference through Monte
Carlo dropout (to sample weights from the Bayesian GCNN). We explored the
performance of the Bayesian GCNN for the task of semi-supervised node
classification and observed that the methodology improved upon
state-of-the-art techniques, particularly for the case where the
number of training labels is small. We also compared the robustness of
Bayesian GCNNs and standard GCNNs under an adversarial attack
involving randomly changing a subset of the edges of node. The
Bayesian GCNN appears to be considerably more resilient to attack.

This paper represents a preliminary investigation into Bayesian graph
convolutional neural networks and focuses on one type of graph model
and one graph learning problem. In future work, we will expand the
approach to other graph models and explore the suitability of the
Bayesian framework for other learning tasks.

\bibliography{references}
\bibliographystyle{aaai}
\end{document}